\documentclass[11pt]{article}
\pdfoutput=1

\usepackage[preprint]{acl}

\usepackage{times}
\usepackage{latexsym}
\usepackage[T1]{fontenc}
\usepackage[utf8]{inputenc}
\usepackage{microtype}
\usepackage{inconsolata}

\usepackage{graphicx}
\graphicspath{{figures/}}
\usepackage{booktabs}
\usepackage{multirow}
\usepackage{colortbl}
\usepackage{xcolor}

\usepackage{amsmath}
\usepackage{amsfonts}
\usepackage{amssymb}
\usepackage{bm}

\usepackage{enumitem}
\usepackage{subcaption}
\usepackage{tikz}
\usetikzlibrary{positioning, arrows.meta, decorations.pathreplacing, calc, fit, backgrounds}

\newcommand{\method}{\textsc{KVSculpt}}
\newcommand{\R}{\mathbb{R}}
\newcommand{\softmax}{\mathrm{softmax}}
\newcommand{\lse}{\mathrm{LSE}}

\title{\method{}: KV Cache Compression as Distillation}

\author{Bo Jiang \\
  Temple University \\
  \texttt{bo.jiang@temple.edu} \And
  Sian Jin \\
  Temple University \\
  \texttt{sian.jin@temple.edu}}

\begin{document}
\maketitle

\begin{abstract}
KV cache compression is critical for efficient long-context LLM inference.
Approaches that reduce the per-pair footprint---quantization and low-rank decomposition---are orthogonal to those that reduce the \emph{sequence length} of the cache.
Along the sequence-length dimension, existing methods range from pure eviction---selecting which KV pairs to keep---to merging, which combines similar pairs into fewer ones.
Both remain anchored to the original cache entries.
We propose \method{}, which moves to the other end of this spectrum: instead of selecting or combining original pairs, we optimize a smaller set of \emph{unconstrained} KV pairs in continuous embedding space to preserve each layer's attention behavior.
Keys are optimized via L-BFGS and values are solved in closed form via least squares, alternating every few steps.
On top of this, we introduce \emph{adaptive budget allocation}, which uses a cheap pilot compression run to redistribute the compression budget across layers and KV heads based on per-component difficulty.

On Qwen2.5-1.5B-Instruct with 2048-token contexts, \method{} reduces KL divergence by $3.5$--$4.1\times$ compared to Select+Fit---attention-score eviction with least-squares value fitting---across compression ratios $r \in \{0.3, 0.5, 0.7\}$.
Adaptive allocation provides an additional $1.3\times$ KL reduction at no extra inference cost.
Analysis reveals that compression difficulty is highly non-uniform: per-layer pilot MSE varies by up to $100\times$ across layers, and the two KV heads within a single layer can differ by up to $467\times$---demonstrating that fine-grained budget allocation is essential.
\end{abstract}

\section{Introduction}
\label{sec:intro}

Autoregressive large language models (LLMs) cache key-value (KV) pairs from all previously generated tokens to avoid recomputation during inference \citep{vaswani2017attention}.
For long contexts, this KV cache becomes the dominant memory bottleneck, consuming tens of gigabytes for a single sequence \citep{kwon2023vllm}.
KV cache compression is therefore essential for practical deployment.

KV cache compression broadly falls into two dimensions: reducing the \emph{size} of each pair (quantization, low-rank) and reducing the \emph{number} of pairs (sequence length).
Along the sequence-length dimension, methods range from pure eviction to merging.
Eviction selects $k$ pairs to keep: criteria include attention score accumulation \citep{zhang2024h2o}, recency with attention sinks \citep{xiao2024efficient}, persistence of importance \citep{liu2024scissorhands}, and pyramidal allocation \citep{cai2024pyramidkv}.
Merging combines similar pairs, modifying values but remaining anchored to the original cache structure.
A hybrid variant fits new values for the selected positions via least squares \citep{devoto2024simple}, but the \emph{keys}---and thus which regions of embedding space are represented---remain constrained to original cache entries.

We argue that this discrete selection framework is unnecessarily restrictive.
After RoPE encoding \citep{su2024roformer}, KV pairs are vectors in a continuous embedding space with no inherent ordering---their positional information is already baked into the embeddings.
There is no reason the compressed cache must be a subset of the original; any set of $k$ vectors that reproduces the correct attention behavior is equally valid.

\begin{figure*}[t]
\centering
\begin{tikzpicture}[
    kvbox/.style={minimum width=0.32cm, minimum height=0.5cm, draw, inner sep=0pt, font=\tiny\sffamily},
    kept/.style={kvbox, fill=blue!25},
    merged/.style={kvbox, fill=violet!25},
    retain/.style={kvbox, fill=green!20},
    free/.style={kvbox, fill=orange!30, rounded corners=2pt},
    origbox/.style={kvbox, fill=gray!15},
    labelstyle/.style={font=\small\sffamily},
    arrowstyle/.style={-{Stealth[length=4pt]}, thick},
    bracecolor/.style={decorate, decoration={brace, amplitude=4pt, mirror}, thick},
]

\def\colA{-5.2}   
\def\colB{0.0}    
\def\colC{5.2}    
\def\bw{0.38}     

\node[labelstyle, font=\sffamily\bfseries\small] at (\colA, 2.5) {(a) Eviction};

\foreach \i/\c in {0/gray!15, 1/gray!15, 2/blue!25, 3/gray!15, 4/blue!25, 5/gray!15, 6/gray!15, 7/blue!25, 8/green!20, 9/green!20, 10/green!20} {
    \node[kvbox, fill=\c] (a\i) at (\colA-1.9+\i*\bw, 1.6) {};
}
\draw[bracecolor] (\colA-1.9-0.14, 1.2) -- node[below=3pt, font=\tiny] {compress zone} (\colA-1.9+7*\bw+0.14, 1.2);
\draw[bracecolor] (\colA-1.9+8*\bw-0.14, 1.2) -- node[below=3pt, font=\tiny] {retain} (\colA-1.9+10*\bw+0.14, 1.2);

\draw[arrowstyle] (\colA, 0.85) -- node[right, font=\tiny, xshift=1pt] {select top-$k$} (\colA, 0.25);

\node[kept] at (\colA-0.76, -0.1) {};
\node[kept] at (\colA-0.76+\bw, -0.1) {};
\node[kept] at (\colA-0.76+2*\bw, -0.1) {};
\node[retain] at (\colA-0.76+3*\bw+0.12, -0.1) {};
\node[retain] at (\colA-0.76+4*\bw+0.12, -0.1) {};
\node[retain] at (\colA-0.76+5*\bw+0.12, -0.1) {};

\node[font=\tiny, text=blue!60!black] at (\colA-0.76+\bw, -0.5) {subset of original};

\node[labelstyle, font=\sffamily\bfseries\small] at (\colB, 2.5) {(b) Merge};

\foreach \i/\c in {0/violet!20, 1/violet!20, 2/gray!15, 3/violet!35, 4/violet!35, 5/violet!35, 6/gray!15, 7/gray!15, 8/green!20, 9/green!20, 10/green!20} {
    \node[kvbox, fill=\c] (b\i) at (\colB-1.9+\i*\bw, 1.6) {};
}
\draw[bracecolor] (\colB-1.9-0.14, 1.2) -- node[below=3pt, font=\tiny] {compress zone} (\colB-1.9+7*\bw+0.14, 1.2);
\draw[bracecolor] (\colB-1.9+8*\bw-0.14, 1.2) -- node[below=3pt, font=\tiny] {retain} (\colB-1.9+10*\bw+0.14, 1.2);

\draw[arrowstyle] (\colB, 0.85) -- node[right, font=\tiny, xshift=1pt, align=left] {group \& avg} (\colB, 0.25);

\node[merged] at (\colB-0.76, -0.1) {};
\node[merged] at (\colB-0.76+\bw, -0.1) {};
\node[merged] at (\colB-0.76+2*\bw, -0.1) {};
\node[retain] at (\colB-0.76+3*\bw+0.12, -0.1) {};
\node[retain] at (\colB-0.76+4*\bw+0.12, -0.1) {};
\node[retain] at (\colB-0.76+5*\bw+0.12, -0.1) {};

\node[font=\tiny, text=violet!60!black] at (\colB-0.76+\bw, -0.5) {avg of neighbors};

\node[labelstyle, font=\sffamily\bfseries\small] at (\colC, 2.5) {(c) \method{} (Ours)};

\foreach \i/\c in {0/gray!15, 1/gray!15, 2/gray!15, 3/gray!15, 4/gray!15, 5/gray!15, 6/gray!15, 7/gray!15, 8/green!20, 9/green!20, 10/green!20} {
    \node[kvbox, fill=\c] (c\i) at (\colC-1.9+\i*\bw, 1.6) {};
}
\draw[bracecolor] (\colC-1.9-0.14, 1.2) -- node[below=3pt, font=\tiny] {compress zone} (\colC-1.9+7*\bw+0.14, 1.2);
\draw[bracecolor] (\colC-1.9+8*\bw-0.14, 1.2) -- node[below=3pt, font=\tiny] {retain} (\colC-1.9+10*\bw+0.14, 1.2);

\draw[arrowstyle] (\colC, 0.85) -- node[right, font=\tiny, xshift=1pt, align=left] {L-BFGS (K)\\lstsq (V)} (\colC, 0.25);

\node[free] at (\colC-0.76, -0.1) {};
\node[free] at (\colC-0.76+\bw, -0.1) {};
\node[free] at (\colC-0.76+2*\bw, -0.1) {};
\node[retain] at (\colC-0.76+3*\bw+0.12, -0.1) {};
\node[retain] at (\colC-0.76+4*\bw+0.12, -0.1) {};
\node[retain] at (\colC-0.76+5*\bw+0.12, -0.1) {};

\node[font=\tiny, text=orange!60!black] at (\colC-0.76+\bw, -0.5) {free in $\mathbb{R}^d$};

\draw[gray, dashed] (-2.6, -0.7) -- (-2.6, 2.7);
\draw[gray, dashed] (2.6, -0.7) -- (2.6, 2.7);

\node[draw, rounded corners=3pt, fill=yellow!8, inner sep=4pt, font=\small, anchor=north] at (0, -0.95) {
    \footnotesize Keys constrained to: \quad
    original positions (a) \quad$\rightarrow$\quad
    combinations of originals (b) \quad$\rightarrow$\quad
    \textbf{unconstrained} $\mathbb{R}^{k \times d}$ (c)
};

\end{tikzpicture}
\caption{Three paradigms for KV cache sequence-length reduction. \textbf{(a)~Eviction} selects a subset of original KV pairs. \textbf{(b)~Merge} combines similar pairs, modifying values but remaining anchored to original positions. \textbf{(c)~\method{}} distills the compress zone into $k$ unconstrained pairs (orange) freely optimized in $\mathbb{R}^d$ via L-BFGS (keys) and least squares (values). All three keep the retain zone (green) intact. The key distinction is the degree of freedom: from discrete subset to weighted combination to fully continuous optimization.}
\label{fig:overview}
\end{figure*}
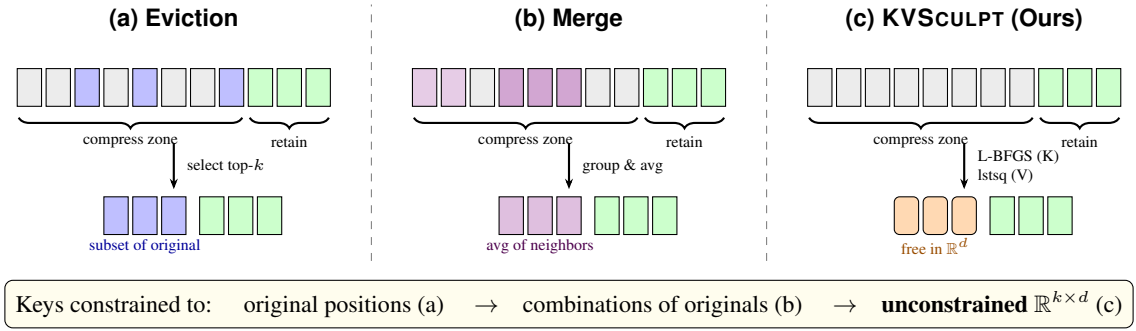

We propose \method{}, which reformulates KV cache compression as \textbf{distillation} \citep{hinton2015distilling} into a smaller cache (Figure~\ref{fig:overview}).
Given a full KV cache of $N$ pairs per layer, we optimize $k$ unconstrained key-value pairs such that the attention output under the compressed cache matches the original.
Keys are optimized with L-BFGS \citep{liu1989lbfgs}, a quasi-Newton method well-suited to the smooth but non-convex attention landscape; values are solved analytically via ridge regression, given the attention weights induced by the current keys.
Optimization is per-layer and per-head, enabling trivial parallelism and bounded memory.

Beyond the core optimizer, we introduce \textbf{adaptive budget allocation}: a short pilot compression run reveals the per-layer and per-head compression difficulty, which is then used to redistribute the fixed total budget.
Layers and heads that are harder to compress receive more pairs; easy ones receive fewer.
At inference time this is free---the same total budget, just redistributed---and the pilot cost is amortized into the compression step.

Our contributions:
\begin{enumerate}[leftmargin=*,itemsep=1pt,topsep=2pt]
  \item We reformulate KV cache compression from discrete selection to distillation, eliminating the combinatorial search over positions.
  \item We propose an L-BFGS + least-squares alternating optimizer that achieves $3.5$--$4.1\times$ lower KL divergence than the best eviction baseline.
  \item We introduce pilot-based adaptive allocation at the layer and head granularity, with per-layer redistribution alone yielding $1.3\times$ lower KL at no extra inference cost.
  \item We provide analysis showing that compression difficulty is highly structured---varying by orders of magnitude across layers and KV heads---and that per-layer errors compound through the transformer, identifying the bottleneck for future work.
\end{enumerate}

\section{Related Work}
\label{sec:related}

\paragraph{Sequence-length reduction.}
Eviction methods select a subset of KV pairs and discard the rest.
H2O \citep{zhang2024h2o} accumulates attention scores across queries and evicts the lowest-scoring pairs.
StreamingLLM \citep{xiao2024efficient} keeps an attention sink window plus recent tokens.
ScissorHands \citep{liu2024scissorhands} exploits the persistence of importance across decoding steps.
PyramidKV \citep{cai2024pyramidkv} allocates different cache sizes per layer based on attention pattern structure.
FastGen \citep{ge2024model} uses attention profiling to decide per-head compression policies.
Rather than discarding pairs entirely, merging methods combine similar ones:
CaM \citep{zhang2024cam} merges low-importance pairs into neighboring important ones via weighted averaging,
D2O \citep{wan2025d2o} distinguishes active and passive tokens and merges the passive ones,
and DMC \citep{nawrot2024dmc} learns a per-head gate that decides whether to append or merge each incoming token.
Both families remain anchored to the original cache entries---eviction keeps a subset unchanged, and merging produces weighted combinations of the originals.

\paragraph{KV cache quantization and low-rank methods.}
Orthogonal lines of work reduce memory by quantizing KV values to lower precision \citep{hooper2024kvquant} or exploiting low-rank structure in the cache.
These approaches are complementary to ours: quantization can be applied on top of the distilled cache, and low-rank compression can be combined with pair reduction.

\paragraph{Adaptive per-layer budgets.}
PyramidKV \citep{cai2024pyramidkv} and PyramidInfer \citep{yang2024pyramidinfer} allocate different cache sizes per layer based on attention entropy.
\citet{devoto2024simple} use L2 norm of values as a compression difficulty signal.
Our work shares the insight that uniform allocation is suboptimal, but differs in signal (dynamic pilot MSE vs.\ static attention patterns) and scope (we also allocate across KV heads within a layer).

\section{Method}
\label{sec:method}

\subsection{Problem Formulation}
\label{sec:formulation}

Consider a single KV head in an attention layer with $h_q$ query heads and $h_\text{kv}$ KV heads, where the head serves $g = h_q / h_\text{kv}$ query heads (GQA group size).
After processing a context of $N$ tokens, the head holds a full KV cache that we partition into a \emph{compress zone} (the oldest $N - m$ pairs) and a \emph{retain zone} (the most recent $m$ pairs, kept unchanged):
\begin{align}
  \mathbf{K}_\text{full} &= [\underbrace{\mathbf{K}_\text{old}}_{N-m} \;;\; \underbrace{\mathbf{K}_\text{ret}}_{m}], &
  \mathbf{V}_\text{full} &= [\underbrace{\mathbf{V}_\text{old}}_{N-m} \;;\; \underbrace{\mathbf{V}_\text{ret}}_{m}]
\end{align}
where $\mathbf{K}_\text{full}, \mathbf{V}_\text{full} \in \R^{N \times d}$ and all keys include RoPE positional encoding.

The goal is to \emph{distill} the compress zone into $k$ freely optimized pairs $(\mathbf{K}_c, \mathbf{V}_c) \in \R^{k \times d}$ such that the compressed cache
\begin{equation}
  \mathbf{K}_\text{cat} = [\mathbf{K}_c \;;\; \mathbf{K}_\text{ret}], \quad
  \mathbf{V}_\text{cat} = [\mathbf{V}_c \;;\; \mathbf{V}_\text{ret}]
\end{equation}
preserves the attention output for any future query.
The compression ratio is $r = (k + m) / N$.

\paragraph{Relationship to eviction and merging.}
Eviction constrains $\mathbf{K}_c$ to a $k$-element subset of rows of $\mathbf{K}_\text{old}$; merging restricts each row to a weighted combination of original rows.
Our feasible set \emph{contains} both: any eviction or merge solution lies in $\R^{k \times d}$, so in principle our optimum cannot be worse.
More broadly, our formulation is a direct answer to the sequence-length reduction problem itself---given a budget of $k$ pairs, find the $k$ pairs in $\R^{k \times d}$ that best preserve attention behavior, with no structural constraints.
Since RoPE already encodes position into the key embedding, $\mathbf{K}_c$ is free to land at ``virtual'' positions that need not correspond to any original token.
In practice, the non-convexity of the softmax landscape means the global optimum is not guaranteed, but L-BFGS with warm-start initialization consistently finds solutions that outperform both eviction and merging (Section~\ref{sec:results}).

\subsection{Loss Function}
\label{sec:loss}

We optimize $\mathbf{K}_c$ and $\mathbf{V}_c$ to match the full-cache attention output for a set of training queries $\mathbf{Q} \in \R^{g \times n_q \times d}$ (construction detailed in Section~\ref{sec:queries}).
Following the chunked attention decomposition used in FlashAttention \citep{dao2022flashattention}, the context chunk's contribution to any future decode step is fully determined by the partial output $\mathbf{o}$ and the log-sum-exp $\ell = \log\sum\exp(\text{scores})$ (which subsumes the max score $\mu$).
As long as these match between the compressed and full cache, the final combined output is identical regardless of the future decode chunk.

This motivates a two-term loss:
\begin{align}
  \mathcal{L} &= \underbrace{\|\hat{\mathbf{Y}} - \mathbf{Y}\|^2_F}_{\text{output MSE}}
  + \underbrace{\|\hat{\bm{\ell}} - \bm{\ell}\|^2_F}_{\text{LSE matching}}
  \label{eq:loss}
\end{align}
where $\mathbf{Y} = \softmax(\mathbf{Q} \mathbf{K}_\text{full}^\top / \sqrt{d})\, \mathbf{V}_\text{full}$ is the full-cache output, $\hat{\mathbf{Y}}$ is the compressed-cache output, and $\bm{\ell} = \lse(\mathbf{Q} \mathbf{K}_\text{full}^\top / \sqrt{d})$ is the log-sum-exp of the full-cache scores (and $\hat{\bm{\ell}}$ the compressed counterpart).
The LSE term ensures that the attention mass assigned to the context chunk is correct, which is critical when the context chunk is later combined with future decode tokens.
We weight both terms equally ($\lambda = 1$); in practice they are comparable in magnitude because both are normalized by the number of queries.

\subsection{Training Query Construction}
\label{sec:queries}

The loss in Eq.~\ref{eq:loss} requires a set of training queries, but the actual future decode queries are unavailable at compression time.
A natural proxy is the \emph{retain queries}---the $m$ real queries from the retain zone---since they are the most recent and thus closest to future decode queries.
However, retain queries carry RoPE at positions $[N{-}m, \ldots, N{-}1]$, while future decode queries will have positions $[N, N{+}1, \ldots]$; this positional mismatch can bias the optimization.

\paragraph{De-RoPE factorization.}
Since $\mathbf{q} = \text{RoPE}(\mathbf{q}_c,\, p)$ where $\mathbf{q}_c = \mathbf{W}_q \mathbf{h}$ is a position-independent content vector, we can recover $\mathbf{q}_c = \text{RoPE}^{-1}(\mathbf{q},\, p)$ by inverting the rotation.
Empirically, $\mathbf{q}_c$ is approximately stationary: consecutive content vectors have cosine similarity $0.91$--$0.93$, and a PCA basis fitted on context tokens captures $81$--$86\%$ of decode variance, with only ${\sim}2\%$ decay per 2048 tokens.
The effective dimensionality is ${\sim}60$ out of 128, indicating a concentrated low-dimensional structure.

\paragraph{Synthetic future queries.}
We uniformly subsample $n_s$ content vectors across the full context, then re-apply RoPE at future positions $N, N{+}1, \ldots, N{+}n_s{-}1$:
\begin{equation}
  \mathbf{Q}_\text{synth} = \text{RoPE}\!\big(\mathbf{q}_c[\text{uniform\_indices}],\; [N, \ldots, N{+}n_s{-}1]\big)
\end{equation}
Uniform sampling provides broad temporal coverage of the content distribution.
We tested alternative strategies (bootstrap from recent tokens, $k$-means centroids, farthest-point sampling, PCA extrapolation with subspace rotation); none improved over uniform, because the distributional drift signals are too small in magnitude to reliably exploit from context alone (Section~\ref{sec:query_ablation}).

\paragraph{Final training set.}
The training queries are the union of all $m$ retain queries (at their original positions) and $n_s$ synthetic future queries, giving $n_q = m + n_s$ total queries per query head ($\mathbf{Q} \in \R^{g \times n_q \times d}$ in Eq.~\ref{eq:loss}).
The retain queries anchor the optimization to real attention patterns, while the synthetic queries improve generalization to future positions.

\subsection{Optimization}
\label{sec:optimization}

The loss in Eq.~\ref{eq:loss} is differentiable w.r.t.\ $\mathbf{K}_c$ (through the softmax) but has a favorable structure for $\mathbf{V}_c$: given fixed attention weights, the optimal $\mathbf{V}_c$ is a linear least-squares solution.

\paragraph{Alternating K-optimization and V-solve.}
We alternate between:
\begin{enumerate}[leftmargin=*,itemsep=0pt]
  \item \textbf{K step}: update $\mathbf{K}_c$ via L-BFGS \citep{liu1989lbfgs} with $\mathbf{V}_c$ frozen.
  L-BFGS uses curvature information from gradient history, which is critical for navigating the non-convex softmax landscape---first-order methods (Adam) get trapped in poor local minima (Section~\ref{sec:lbfgs_vs_adam}).
  \item \textbf{V step} (every 5 K steps): solve $\mathbf{V}_c^* = \arg\min_{\mathbf{V}_c} \|\mathbf{A}_c \mathbf{V}_c + \mathbf{A}_r \mathbf{V}_\text{ret} - \mathbf{Y}\|^2_F + \lambda_r\|\mathbf{V}_c\|^2_F$ via ridge regression ($\lambda_r = 10^{-3}$), where $\mathbf{A} = \softmax(\mathbf{Q}\mathbf{K}_\text{cat}^\top/\sqrt{d}) \in \R^{n_q \times (k+m)}$ is partitioned as $[\mathbf{A}_c \;;\; \mathbf{A}_r]$ over the compressed and retained key positions.
\end{enumerate}

\paragraph{Initialization.}
We initialize $\mathbf{K}_c$ from the top-$k$ positions by attention importance score (sum of softmax attention weights across all queries), which provides a warm start near a good basin.

\paragraph{Per-head independence.}
With grouped-query attention \citep{ainslie2023gqa}, each KV head serves a group of query heads independently.
The loss decomposes across KV heads, so we optimize each head separately.
This enables per-head budget allocation (Section~\ref{sec:adaptive}).

\subsection{Adaptive Budget Allocation}
\label{sec:adaptive}

The standard approach allocates the same number of compressed pairs $k$ to every layer and head.
However, compression difficulty varies dramatically across components.

\paragraph{Pilot-MSE signal.}
We run a short pilot compression with uniform allocation to obtain a per-component MSE signal (60 L-BFGS steps for per-layer allocation; 30 steps per head for per-head allocation).
This signal captures the \emph{sequence-specific} compression difficulty of each layer and head---unlike static signals such as value norm or attention entropy, which are model properties independent of the input.

\paragraph{Two-level allocation.}
Given a fixed total budget $B = k \times L \times h_\text{kv}$ (where $L$ is the number of layers):
\begin{enumerate}[leftmargin=*,itemsep=0pt]
  \item \textbf{Per-layer}: compute mean pilot MSE across heads for each layer; allocate layer budgets proportional to $\text{MSE}_l^{0.5}$ (square-root dampening prevents outlier layers from consuming the entire budget).
  \item \textbf{Per-head within layer}: given the layer budget, redistribute across KV heads proportional to $\text{MSE}_{l,h}^{0.5}$.
\end{enumerate}

The square-root dampening is critical: undampened MSE ($\alpha = 1.0$) causes over-allocation to outlier layers (e.g., Layer~0, whose MSE is $10$--$100\times$ the median), starving other layers and worsening overall quality.

\section{Experimental Setup}
\label{sec:experiments}

\paragraph{Model and data.}
We use Qwen2.5-1.5B-Instruct \citep{qwen2025qwen25}, a 28-layer model with grouped-query attention (12 query heads, 2 KV heads, head dimension 128).
We sample sequences from the PG19 test set with context length $N = 2048$ and evaluate on 128 continuation tokens.
Baseline experiments use 20 sequences; detailed analysis and ablations use a 5-sequence subset.

\paragraph{Evaluation metric.}
We measure KL divergence between the output logits under compressed KV and the ground-truth logits under the full cache, computed over the 128 continuation tokens with teacher forcing.
Lower KL indicates better preservation of the model's output distribution.

\paragraph{Compression ratios.}
We test $r \in \{0.1, 0.2, 0.3, 0.5, 0.7\}$.
The retain zone is $m = 256$ tokens (the most recent 12.5\% of the context).

\paragraph{Baselines.}
We compare against four methods:
\begin{itemize}[leftmargin=*,itemsep=1pt]
  \item \textbf{Random}: randomly select $k$ positions to keep.
  \item \textbf{Attention Score}: keep the top-$k$ positions by accumulated attention score across all queries \citep{zhang2024h2o}.
  \item \textbf{Select+Fit}: select positions by attention score, then fit values via ridge regression. Inspired by \citet{devoto2024simple}, who use $L_2$ norm for selection; we substitute attention-score selection for a stronger baseline.
  \item \textbf{Joint Optimization}: learn binary gates via Hard Concrete relaxation \citep{louizos2018learning} jointly with KV values; the discrete selection analogue of our method.
\end{itemize}

\paragraph{\method{} configuration.}
L-BFGS with learning rate 0.5, strong Wolfe line search, 10 inner iterations per step; 100 outer steps per layer; V solved via ridge regression ($\lambda = 10^{-3}$) every 5 steps; 128 synthetic future queries; retain zone $m = 256$.

\section{Results}
\label{sec:results}

\subsection{Main Results: Distillation vs.\ Eviction}
\label{sec:main_results}

Table~\ref{tab:main} compares \method{} against baselines across compression ratios.

\begin{table}[t]
\centering
\small
\caption{KL divergence ($\downarrow$) on 5 sequences (mean). \method{} consistently outperforms all eviction baselines by a wide margin.}
\label{tab:main}
\begin{tabular}{lccc}
\toprule
\textbf{Method} & $r{=}0.3$ & $r{=}0.5$ & $r{=}0.7$ \\
\midrule
Random         & $2.50$          & $1.80$          & $1.54$          \\
Attn Score     & $2.54\mathrm{e}{\text{-}1}$  & $2.04\mathrm{e}{\text{-}1}$ & $1.39\mathrm{e}{\text{-}1}$ \\
Select+Fit     & $2.33\mathrm{e}{\text{-}1}$  & $1.86\mathrm{e}{\text{-}1}$ & $1.25\mathrm{e}{\text{-}1}$ \\
Joint Opt      & $2.24\mathrm{e}{\text{-}1}$  & $1.80\mathrm{e}{\text{-}1}$ & $1.15\mathrm{e}{\text{-}1}$ \\
\midrule
\textbf{\method{}} & $\mathbf{5.75\mathrm{e}{\text{-}2}}$ & $\mathbf{4.63\mathrm{e}{\text{-}2}}$ & $\mathbf{3.58\mathrm{e}{\text{-}2}}$ \\
\midrule
\textit{vs.\ Select+Fit} & $4.1\times$ & $4.0\times$ & $3.5\times$ \\
\bottomrule
\end{tabular}
\end{table}

\method{} achieves $3.5$--$4.1\times$ lower KL than Select+Fit, the strongest eviction baseline, across all tested ratios.
The advantage is most pronounced at aggressive compression ($r = 0.3$, $4.1\times$) where the discrete selection problem is hardest.

\paragraph{The role of continuous keys.}
Joint Optimization uses the same per-layer MSE objective as \method{} and also optimizes values, but constrains keys to original cache positions via Hard Concrete gates.
It barely improves over Select+Fit ($2.24$ vs.\ $2.33\mathrm{e}{\text{-}1}$ at $r{=}0.3$), showing that \emph{optimizing values alone is insufficient}---the key advantage of \method{} is moving keys freely in $\R^d$, not just fitting better values.

\paragraph{Per-sequence variation.}
We label sequences with \method{} KL $< 0.01$ at $r{=}0.3$ as \emph{easy} (3 of 5) and the rest as \emph{hard}.
Select+Fit remains at KL $> 0.1$ for all sequences, including the easy ones.
On harder sequences (Seq~1, Seq~2), the advantage narrows to $1.2$--$1.8\times$ but never reverses (Table~\ref{tab:per_seq}).

\begin{table}[t]
\centering
\small
\caption{Per-sequence KL at $r{=}0.3$. \method{} achieves near-lossless compression on easy sequences and remains superior on hard ones.}
\label{tab:per_seq}
\begin{tabular}{lccc}
\toprule
\textbf{Seq} & \textbf{Select+Fit} & \textbf{\method{}} & \textbf{Improvement} \\
\midrule
0 (easy) & $1.28\mathrm{e}{\text{-}1}$ & $8.1\mathrm{e}{\text{-}3}$  & $15.8\times$ \\
1 (hard) & $1.39\mathrm{e}{\text{-}1}$ & $1.17\mathrm{e}{\text{-}1}$ & $1.2\times$  \\
2 (hard) & $2.61\mathrm{e}{\text{-}1}$ & $1.43\mathrm{e}{\text{-}1}$ & $1.8\times$  \\
3 (easy) & $3.92\mathrm{e}{\text{-}1}$ & $9.7\mathrm{e}{\text{-}3}$  & $40.5\times$ \\
4 (easy) & $2.46\mathrm{e}{\text{-}1}$ & $9.6\mathrm{e}{\text{-}3}$  & $25.5\times$ \\
\midrule
Mean     & $2.33\mathrm{e}{\text{-}1}$ & $5.75\mathrm{e}{\text{-}2}$ & $4.1\times$  \\
\bottomrule
\end{tabular}
\end{table}

\subsection{L-BFGS vs.\ Adam and Near-Optimality}
\label{sec:lbfgs_vs_adam}

The choice of optimizer is critical.
Replacing L-BFGS with Adam (the standard first-order optimizer for neural network training) degrades per-layer MSE by $17$--$95\times$ (Table~\ref{tab:optimizer}).
This translates to $8$--$15\times$ worse end-to-end KL on easy sequences.

The softmax in the attention computation creates a loss landscape with sharp, narrow valleys.
L-BFGS's curvature approximation navigates these efficiently, while Adam's diagonal preconditioning is insufficient.

\paragraph{Near-optimality.}
To assess how close L-BFGS gets to the best achievable per-layer solution, we run an oracle search: 100 random restarts with the best result taken as an empirical upper bound.
Table~\ref{tab:optimizer} shows that a single L-BFGS run (1 restart) already reaches within 2--8\% of the 100-restart oracle.
This near-optimality has an important implication: \emph{the per-layer optimization problem is essentially solved}.
The remaining gap to perfect end-to-end quality is not due to suboptimal per-layer compression, but due to error accumulation across layers---a limitation of the per-layer objective itself (Section~\ref{sec:analysis}).

\begin{table}[t]
\centering
\small
\caption{Per-layer MSE ($r{=}0.3$). L-BFGS (1 restart) is $17$--$95\times$ better than Adam and within 2--8\% of the 100-restart oracle, indicating near-optimal per-layer solutions.}
\label{tab:optimizer}
\begin{tabular}{lcccc}
\toprule
\textbf{Layer} & \textbf{Adam} & \textbf{L-BFGS} & \textbf{Oracle} & \textbf{Gap} \\
\midrule
L2   & $1.78\mathrm{e}{\text{-}3}$ & $1.05\mathrm{e}{\text{-}4}$ & $9.67\mathrm{e}{\text{-}5}$ & $8\%$ \\
L14  & $7.97\mathrm{e}{\text{-}2}$ & $1.28\mathrm{e}{\text{-}3}$ & $1.20\mathrm{e}{\text{-}3}$ & $7\%$ \\
L26  & $1.14\mathrm{e}{\text{-}1}$ & $1.20\mathrm{e}{\text{-}3}$ & $1.18\mathrm{e}{\text{-}3}$ & $2\%$ \\
\bottomrule
\end{tabular}
\end{table}

\subsection{Query Strategy Ablation}
\label{sec:query_ablation}

We compare five strategies for constructing the synthetic future queries described in Section~\ref{sec:queries}, evaluated on attention cosine similarity at near (128 tokens) and far (4096 tokens) horizons (Table~\ref{tab:query_ablation}).

\begin{table}[t]
\centering
\small
\caption{Query sampling strategies. Uniform spread provides the best overall trade-off, especially at far horizons.}
\label{tab:query_ablation}
\begin{tabular}{lcc}
\toprule
\textbf{Strategy} & \textbf{Near attn cos} & \textbf{Far attn cos} \\
\midrule
Bootstrap (last 128) & \textbf{0.950} & 0.737 \\
Uniform spread       & 0.948 & \textbf{0.759} \\
Random sample        & 0.948 & 0.756 \\
$k$-means centroids  & 0.947 & 0.756 \\
Farthest-point       & 0.925 & 0.758 \\
\bottomrule
\end{tabular}
\end{table}

Bootstrap (sampling recent tokens only) wins at the near horizon but degrades at far horizons because the content vector distribution shifts slightly over time.
Uniform spread sacrifices $0.2\%$ near-term accuracy for $3\%$ far-term improvement by covering the full temporal span of the context.
Advanced strategies (PCA extrapolation, subspace rotation prediction, norm correction) were tested but none improved over uniform---the distributional drift signals (${\sim}2\%$ subspace rotation per 2048 tokens, $5\%$ norm growth) are too small to reliably exploit.

\subsection{Adaptive Budget Allocation}
\label{sec:adaptive_results}

The preceding sections establish the core optimizer; we now turn to \emph{how to distribute the compression budget}.
Since optimization is per-head, we can measure each component's difficulty and allocate accordingly.

\paragraph{Static signals fail across sequences.}
We first test allocation based on value norm ($\|\mathbf{V}\|$), a static model property.
While $\|\mathbf{V}\|^{0.5}$-weighted allocation improves Seq~1 by 25\%, it \emph{worsens} Seq~2 by 73\% (Table~\ref{tab:vnorm}).
The mean KL across 5 sequences is 27\% worse than uniform.
The problem is that $\|\mathbf{V}\|$ is a model property invariant across sequences, but compression difficulty is sequence-dependent.

\begin{table}[t]
\centering
\small
\caption{Static allocation ($\|\mathbf{V}\|^{0.5}$) is unstable across sequences. Multipliers are relative to uniform allocation ($r{=}0.3$).}
\label{tab:vnorm}
\begin{tabular}{lcc}
\toprule
\textbf{Seq} & \textbf{Uniform KL} & $\|\mathbf{V}\|^{0.5}$ \textbf{vs.\ Uni.} \\
\midrule
0 & $8.1\mathrm{e}{\text{-}3}$ & $1.05\times$  \\
1 & $1.17\mathrm{e}{\text{-}1}$ & $\mathbf{0.75\times}$  \\
2 & $1.43\mathrm{e}{\text{-}1}$ & $1.73\times$  \\
3 & $9.7\mathrm{e}{\text{-}3}$ & $0.84\times$  \\
4 & $9.6\mathrm{e}{\text{-}3}$ & $1.23\times$  \\
\midrule
Mean & $5.75\mathrm{e}{\text{-}2}$ & $1.27\times$ \\
\bottomrule
\end{tabular}
\end{table}

\paragraph{Pilot-MSE allocation.}
A 60-step pilot run at uniform allocation produces a per-layer MSE signal that is \emph{sequence-specific}.
Allocating proportional to $\text{MSE}^{0.5}$ wins on 4/5 sequences and reduces mean KL by 25\% (Table~\ref{tab:pilot}).

\begin{table}[t]
\centering
\small
\caption{Pilot-MSE$^{0.5}$ allocation wins 4/5 sequences with 25\% mean KL reduction ($r{=}0.3$).}
\label{tab:pilot}
\begin{tabular}{lccc}
\toprule
\textbf{Seq} & \textbf{Uniform} & \textbf{Pilot-MSE$^{0.5}$} & \textbf{vs.\ Uniform} \\
\midrule
0 & $8.14\mathrm{e}{\text{-}3}$ & $8.93\mathrm{e}{\text{-}3}$  & $1.10\times$ \\
1 & $1.170\mathrm{e}{\text{-}1}$ & $7.01\mathrm{e}{\text{-}2}$ & $\mathbf{0.60\times}$ \\
2 & $1.433\mathrm{e}{\text{-}1}$ & $1.208\mathrm{e}{\text{-}1}$ & $\mathbf{0.84\times}$ \\
3 & $9.67\mathrm{e}{\text{-}3}$ & $8.64\mathrm{e}{\text{-}3}$  & $\mathbf{0.89\times}$ \\
4 & $9.59\mathrm{e}{\text{-}3}$ & $7.77\mathrm{e}{\text{-}3}$  & $\mathbf{0.81\times}$ \\
\midrule
Mean & $5.75\mathrm{e}{\text{-}2}$ & $4.31\mathrm{e}{\text{-}2}$ & $\mathbf{0.75\times}$ \\
\bottomrule
\end{tabular}
\end{table}

The improvement is largest on hard sequences (Seq~1: $0.60\times$) and slightly negative on the easiest sequence (Seq~0: $1.10\times$), where the baseline KL is already near zero and reallocation adds noise.

\paragraph{Square-root dampening.}
Undampened allocation ($\text{MSE}^{1.0}$) is too aggressive: it gives Layer~0 (which has $10$--$100\times$ the median MSE) an extreme share of the budget, starving other layers.
At $\alpha = 1.0$, mean KL is only $0.97\times$ uniform (barely improved), while $\alpha = 0.5$ achieves $0.75\times$.

\subsection{Per-Head Allocation}
\label{sec:perhead_results}

Per-layer allocation treats each layer as a unit, but with grouped-query attention, the model has 2 KV heads per layer.
These heads serve qualitatively different functions: Layer~15 is consistently the most asymmetric across all sequences (17--56$\times$ head MSE ratio), with Head~0 easy to compress and Head~1 hard; Layer~0 shows the opposite pattern with even larger ratios (up to $467\times$).
This structural asymmetry persists across all tested sequences, motivating per-head budget redistribution.

Quantitatively, the per-head pilot MSE ratio averages $3.7$--$21.5\times$ across layers (Table~\ref{tab:head_asym}).

\begin{table}[t]
\centering
\small
\caption{Head MSE asymmetry: mean and max ratio of per-head pilot MSE within the same layer ($r{=}0.3$).}
\label{tab:head_asym}
\begin{tabular}{lccc}
\toprule
\textbf{Seq} & \textbf{Mean ratio} & \textbf{Max ratio} & \textbf{Layer} \\
\midrule
0 & $21.5\times$ & $467\times$ & L0  \\
1 & $5.5\times$  & $31\times$  & L15 \\
2 & $6.1\times$  & $33\times$  & L15 \\
3 & $3.7\times$  & $17\times$  & L15 \\
4 & $6.2\times$  & $56\times$  & L15 \\
\bottomrule
\end{tabular}
\end{table}

Table~\ref{tab:perhead} isolates the effect of per-head redistribution.
Here, ``per-layer'' uses a 30-step per-head pilot averaged to layer level (noisier than the 60-step joint pilot in Table~\ref{tab:pilot}, hence the weaker per-layer numbers).
Adding per-head allocation on top consistently improves KL, by 1--58\% depending on the sequence; the per-layer+per-head strategy wins 4 of 5 sequences.

\begin{table}[t]
\centering
\small
\caption{Per-head allocation always improves over per-layer alone. All numbers relative to uniform ($r{=}0.3$).}
\label{tab:perhead}
\begin{tabular}{lccc}
\toprule
\textbf{Seq} & \textbf{Per-layer} & \textbf{Per-layer+head} & $\Delta$ \\
\midrule
0 & $0.82\times$ & $\mathbf{0.81\times}$ & $-1\%$ \\
1 & $1.37\times$ & $\mathbf{1.15\times}$ & $-16\%$ \\
2 & $0.88\times$ & $\mathbf{0.84\times}$ & $-5\%$ \\
3 & $0.87\times$ & $\mathbf{0.86\times}$ & $-1\%$ \\
4 & $1.77\times$ & $\mathbf{0.75\times}$ & $-58\%$ \\
\midrule
Mean & $1.05\times$ & $\mathbf{0.92\times}$ & $-8\%$ \\
\bottomrule
\end{tabular}
\end{table}

\section{Analysis: Limits of Per-Layer Optimization}
\label{sec:analysis}

The results above show that \method{} nearly solves the per-layer optimization problem (within 2--8\% of the oracle).
Yet hard sequences still exhibit high KL.
We now investigate why: the bottleneck is not per-layer quality, but how errors propagate across layers and concentrate in specific tokens.

\subsection{Error Accumulation Across Layers}

Per-layer optimization does not guarantee end-to-end quality because errors compound through the transformer.
We measure hidden state MSE at each layer during decoding with compressed KV and find that errors compound by two to three orders of magnitude from the first to the last layer ($220\times$ for easy sequences, up to $5800\times$ for hard ones).

Crucially, the per-layer compression MSE and end-to-end KL are only weakly correlated.
Sequence~0 has $2.4\times$ higher per-layer MSE than Sequence~1, yet Sequence~1 has $14\times$ higher end-to-end KL.
Some sequences are structurally more sensitive to small perturbations---an effect invisible to the per-layer objective.

\subsection{Per-Token KL Concentration}

KL divergence is not spread uniformly across continuation tokens.
On the hardest sequence at $r = 0.3$, 82\% of the total KL is concentrated in just 5 of 128 tokens.
The maximum per-token KL is $7.17$, while the mean is $0.117$ (a $61\times$ ratio).
These ``sensitive'' tokens tend to occur at high-entropy decision points (e.g., first token of a new clause), where the softmax distribution is flat and small logit perturbations cause large probability shifts.

\section{Discussion}
\label{sec:discussion}

\paragraph{Why does continuous optimization help so much?}
As shown in Section~\ref{sec:main_results}, optimizing values while keeping keys at original positions (Joint Optimization) barely helps.
The decisive factor is key freedom: a distilled key can ``summarize'' multiple original keys by positioning itself in embedding space such that its attention pattern approximates their combined effect---a representation that no single original token contains.
This is why the eviction-to-distillation gap ($4\times$) far exceeds the eviction-to-joint-optimization gap ($1.04\times$).

\paragraph{Deployment scenario.}
\method{} is designed for the \emph{offline} setting: given a long context that will be queried many times (e.g., a document, a system prompt, a retrieved passage), the compression cost is amortized over many decode steps.
At 100 L-BFGS steps per layer, compression takes $\sim$170s for a 2048-token context on a single A100 GPU.

\paragraph{The next bottleneck.}
Our analysis shows that the per-layer problem is essentially solved (within 2--8\% of the oracle), yet hard sequences remain far from lossless.
The bottleneck is cross-layer error propagation: small per-layer perturbations compound by 2--3 orders of magnitude.
Cascade-aware optimization---where each layer's target accounts for upstream compression error---is a natural next step, and the per-head decomposition makes this tractable.

\section{Conclusion}

We introduced \method{}, which reframes KV cache compression from discrete token selection to continuous distillation.
The key insight is that after RoPE encoding, KV pairs are free vectors in $\R^d$---there is no reason the compressed cache must be a subset of the original.
By optimizing keys with L-BFGS and solving values in closed form, \method{} achieves near-lossless compression on easy sequences (KL $< 0.01$) at ratios where eviction baselines remain an order of magnitude worse.
Adaptive budget allocation, guided by a cheap pilot signal, provides a further $1.3\times$ reduction by exploiting the extreme non-uniformity of compression difficulty across layers and heads.
Our analysis identifies cross-layer error propagation as the remaining bottleneck, suggesting cascade-aware optimization as a promising direction.

\section*{Limitations}

\begin{itemize}[leftmargin=*,itemsep=2pt]
  \item \textbf{Single model}: All experiments use Qwen2.5-1.5B-Instruct. Validation on other architectures (Llama, Mistral) and scales (7B+) is needed.
  \item \textbf{Fixed context length}: We test only $N = 2048$. Long-context scenarios ($8$K--$128$K) may exhibit different compression dynamics.
  \item \textbf{Compression cost}: At $\sim$170s per context on an A100, \method{} is suitable for offline/amortized settings but not for online single-pass decoding.
  \item \textbf{KL-only evaluation}: We measure logits KL divergence but do not evaluate downstream task accuracy (e.g., MMLU, summarization quality).
  \item \textbf{No merge baselines}: We compare against eviction methods but not merging approaches (CaM, D2O, DMC). While our formulation theoretically subsumes merging, empirical comparison would strengthen the claim.
  \item \textbf{No comparison with quantization}: KV quantization \citep{hooper2024kvquant} is complementary and could be combined with \method{}, but this is not explored.
  \item \textbf{Per-layer optimization}: Errors accumulate across layers in ways not captured by the per-layer objective. Global or cascade-aware optimization could improve hard sequences.
\end{itemize}

\appendix

\section{Full Baseline Results (20 Sequences)}
\label{sec:appendix_full}

Table~\ref{tab:full_baseline} reports KL divergence across all 20 sequences and 5 compression ratios for the baseline methods.

\begin{table}[h]
\centering
\small
\caption{KL divergence (mean $\pm$ std, 20 sequences) at representative ratios. Joint Opt marginally improves over Select+Fit; both are far behind \method{}.}
\label{tab:full_baseline}
\setlength{\tabcolsep}{4pt}
\resizebox{\columnwidth}{!}{%
\begin{tabular}{lcccc}
\toprule
\textbf{Method} & $r{=}0.1$ & $r{=}0.3$ & $r{=}0.5$ & $r{=}0.7$ \\
\midrule
Random     & $3.43{\pm}1.21$ & $2.60{\pm}0.45$ & $2.06{\pm}0.62$ & $1.83{\pm}0.57$ \\
Attn Score & $.322{\pm}.105$ & $.198{\pm}.084$ & $.137{\pm}.065$ & $.086{\pm}.056$ \\
Select+Fit & $.322{\pm}.107$ & $.185{\pm}.076$ & $.129{\pm}.057$ & $.080{\pm}.050$ \\
Joint Opt  & $.311{\pm}.113$ & $.182{\pm}.073$ & $.121{\pm}.058$ & $.073{\pm}.047$ \\
\bottomrule
\end{tabular}%
}
\end{table}

\section{Dampening Exponent Sensitivity}
\label{sec:appendix_exponent}

Table~\ref{tab:exponent} shows the effect of the dampening exponent $\alpha$ in $\|\mathbf{V}\|^\alpha$-weighted allocation on Seq~1 ($r{=}0.3$).
The relationship is non-monotonic: $\alpha = 0.5$ is optimal.
Pilot-MSE allocation shows the same dampening pattern ($0.97\times$ at $\alpha{=}1.0$, $0.75\times$ at $\alpha{=}0.5$, mean over 5 sequences).

\begin{table}[h]
\centering
\small
\caption{Dampening exponent $\alpha$ for $\|\mathbf{V}\|^\alpha$ per-layer allocation (Seq~1, $r{=}0.3$). $\alpha{=}0.5$ is optimal; higher values over-allocate to outlier layers.}
\label{tab:exponent}
\begin{tabular}{lcccc}
\toprule
$\alpha$ & \textbf{KL} & \textbf{vs.\ Uniform} & $k_\text{min}$ & $k_\text{max}$ \\
\midrule
0 (uniform) & $1.170\mathrm{e}{\text{-}1}$ & ref & 358 & 358 \\
0.3 & $1.175\mathrm{e}{\text{-}1}$ & $1.00\times$ & 277 & 480 \\
\textbf{0.5} & $\mathbf{8.83\mathrm{e}{\text{-}2}}$ & $\mathbf{0.75\times}$ & 231 & 578 \\
0.7 & $1.328\mathrm{e}{\text{-}1}$ & $1.14\times$ & 191 & 689 \\
1.0 & $9.27\mathrm{e}{\text{-}2}$ & $0.79\times$ & 140 & 880 \\
\bottomrule
\end{tabular}
\end{table}

\end{document}